\documentclass[a4paper,fleqn，11pt]{cas-dc}

\usepackage[numbers]{natbib}

\def\tsc#1{\csdef{#1}{\textsc{\lowercase{#1}}\xspace}}
\tsc{WGM}
\tsc{QE}
\tsc{EP}
\tsc{PMS}
\tsc{DE}

\begin{document}
\let\WriteBookmarks\relax
\def\floatpagepagefraction{1}
\def\textpagefraction{.001}
\let\printorcid\relax
\shorttitle{Luminance-Aware Multi-Scale Polarization Image Fusion: A Framework with a New Dataset for Complex Lighting}

\title [mode = title]{Luminance-Aware Multi-Scale Polarization Image Fusion: A Framework with a New Dataset for Complex Lighting}

\author[author1]{Zhuangfan~Huang}
\ead{2112455033@stu.fosu.edu.cn}

\author[author1]{Xiaosong~Li}
\ead{lixiaosong@buaa.edu.cn}
\cormark[1]

\author[label1]{Gao~Wang}
\ead{wanggao@nuc.edu.cn}

\author[author1]{Yang~Liu}
\ead{ly25@fosu.edu.cn}

\author[label2]{Tao~Ye}
\ead{ayetao198715@163.com}

\author[author1]{Haishu~Tan}
\ead{tanhaishu@fosu.edu.cn}

\author[label3]{Huafeng~Li}
\ead{lhfchina99@kust.edu.cn}


\address[author1]{Guangdong-HongKong-Macao Joint Laboratory for Intelligent Micro-Nano Optoelectronic Technology, School of Physics and Optoelectronic Engineering, Foshan University, Foshan 528225, China}

\address[label1]{State Key Laboratory of Dynamic Measurement Technology, North University of China, Taiyuan 030051, China}

\address[label2]{School of Mechanicaland Electrical Engineering, China University of Mining and Technology(Beijing), Beijing 100083, China.}

\address[label3]{School of Information Engineering and Automation, Kunming University of Science and Technology, Kunming 650500,China}

\begin{abstract}
Polarization image fusion integrates images$S_{0}$ and $DoLP$ to enhance surface texture and material properties, empowering breakthroughs in camouflage recognition, tissue pathology analysis, and surface defect detection. However, existing methods fail to simultaneously retain $S_{0}$ texture and $DoLP$ contrast in complex lighting, producing halos or loss of features. In this work, we develop a luminance-aware multi-scale polarization image fusion framework to explicitly address the mismatch between texture preservation and polarization-sensitive contrast enhancement under complex illumination. The proposed network incorporates a Texture-Fusion Block (TFB) to capture complementary multi-scale texture features from $S_{0}$ and $DoLP$, and a Hierarchical Luminance Prior Generator (HLPG) to inject hierarchical luminance priors into the encoder for luminance-guided feature enhancement. In addition, a global-local interaction mechanism based on lightweight Swin attention and CBAM is employed to couple broad contextual relationships with local structural refinement. During reconstruction, an Adaptive Photometric Rectification (APR) module is further introduced to perform luminance-adaptive correction by learning the mapping between luminance distribution and fused texture representation. To support evaluation under realistic lighting variation, we also construct MSP, a new multi-scene polarization image fusion benchmark containing 1,000 image pairs acquired from 17 indoor and outdoor complex lighting scenarios. Extensive experiments on MSP, PIF and GAND datasets verify that the proposed method achieves consistently competitive performance in both subjective and objective evaluations, and the MS-SSIM and SD metrics are higher than the average values of other methods by 8.57\%, 60.64\%, 10.26\%, 63.53\%, 22.21\%, and 54.31\%, respectively. The source code and datasets are available at \href{https://github.com/1hzf/MLS-UNet}{https://github.com/1hzf/MLSN}
\end{abstract}


\maketitle

\section{Introduction}
Polarization is a fundamental vector property of light that provides complementary information beyond conventional intensity imaging. By characterizing the change of polarization states after light interacts with object surfaces, polarization imaging can reveal material-dependent cues such as surface roughness, geometric structure, and reflectance behavior. In practical systems, polarization information is commonly represented by the Stokes formalism \cite{1}, from which the total intensity image$S_{0}$, the degree of linear polarization ($DoLP$) and the angle of polarization ($AoP$) can be derived. Owing to its unique ability to capture material- and structure-related cues, polarization image fusion has attracted increasing attention in applications such as security inspection, underwater vision, dehazing, and camouflage perception.

Early polarization image fusion methods mainly relied on mathematical transformations and hand-crafted fusion rules. These conventional approaches can be broadly categorized into three groups: spatial domain methods\cite{6}, transform domain methods\cite{8}, and sparse representation-based methods\cite{10}. Although they provide certain physical interpretability, they still suffer from limited global consistency, weak scene adaptability, and relatively high computational cost, which restrict their applicability in complex real-world environments \cite{13}.

With the development of deep learning, a variety of neural architectures have been introduced into polarization image fusion. Compared with traditional approaches, these methods improve fusion performance through adaptive feature extraction, physical-prior-guided modeling, and end-to-end optimization \cite{14}. Existing deep learning-based methods can be broadly grouped into CNN-based \cite{18}, Transformer-based \cite{19,20}, Mamba-based \cite{25}, diffusion-based \cite{27}, and GAN-based \cite{32} frameworks. Despite their progress, three limitations remain prominent. First, most methods do not explicitly model illumination variation during fusion, which often leads to detail loss in shadowed or highly reflective regions. Second, the polarization-sensitive cues contained in $DoLP$ are still insufficiently exploited, weakening material- and structure-aware representation. Third, publicly available datasets still provide limited coverage of polarization-responsive materials and illumination conditions, making it difficult to learn robust fusion models for realistic scenes.

To address the above limitations, we propose a luminance-aware multi-scale polarization image fusion network. The proposed framework is built upon three key components. First, the TFB is designed to extract complementary multi-scale texture representations from $S_{0}$ and $DoLP$. Second, the HLPG is designed to generate multi-level luminance priors that guide feature modulation throughout the network, thereby improving detail preservation under complex illumination. Third, a global-local interaction scheme based on lightweight Swin attention and CBAM is employed to enhance contextual modeling and structural refinement. In addition, the APR module is used during reconstruction to establish an explicit mapping between luminance distribution and fused texture features, enabling adaptive correction of the final output. Through this design, the proposed framework achieves a better balance between structural fidelity, luminance consistency, and robustness in complex lighting conditions.
The main contributions of this paper can be summarized as follows:
\relpenalty=10000
\binoppenalty=10000
\sloppy
\begin{enumerate}
\itemsep=0pt
\item We propose a luminance-aware multi-scale polarization image fusion network that explicitly models illumination variation during fusion. The proposed framework integrates TFB, HLPG and APR to improve the preservation of polarization-sensitive details under complex lighting conditions.
\item We design a multi-objective loss function tailored for polarization image fusion, which jointly constrains structural similarity, pixel-level fidelity, texture preservation, and contrast consistency to stabilize network optimization under complex illumination conditions.
\item We construct a new multi-scene polarization image fusion benchmark, MSP, which contains 1,000 samples with four-direction polarization observations and derived polarization components, covering diverse materials and illumination scenarios. Extensive experiments on MSP, PIF, and GAND demonstrate that the proposed method achieves consistently competitive performance, with particularly clear advantages on the proposed MSP benchmark.
\end{enumerate}  

The rest of the paper is structured as follows: in Section II, we provide an overview of the development of relevant polarization image fusion. In Section III, we describe the proposed network framework in detail. In Section IV, quantitative experiments on the proposed method are presented. Finally, Section V provides a comprehensive summary of the article.
\section{Related works}

\subsection{Derivation of $DoLP$ and $S_{0}$}
Stokes vector representation: Polarized light is one of the fundamental properties of light and is highly sensitive to the microstructure and optical properties of objects, which makes it an important cue in optical sensing. Compared with the Jones vector formalism, the Stokes vector representation, originally proposed by G. G. Stokes, can describe partially polarized light as well as natural light using a four-dimensional vector. In addition, the Stokes vector can be obtained by combining the polarized components of a light beam measured at different orientations. Accordingly, the Stokes vector $S$ is defined as follows:

\begin{equation}
\label{eq1}
\hspace{+8mm}
S  =  \left[\begin{array}{l}
S_{0} \\
S_{1} \\
S_{2} \\
S_{3}
\end{array}\right]  
= \left[\begin{array}{c}
I_{H}+I_{V} \\
I_{H}-I_{V} \\
I_{45}-I_{135} \\
I_{R}-I_{L}
\end{array}\right]
\end{equation}
In Eq(\ref{eq1}), $I_{H}$ and $I_{V}$ denote the intensities of linearly polarized light along the horizontal and vertical directions, respectively; $I_{45}$ and $I_{135}$ denote the intensities measured at polarization angles of 45° and 135°, respectively; $I_{R}$ and $I_{L}$ denote the intensities of circularly polarized light from the right and the left, respectively. Similarly, $S_{0}$ represents the total intensity of the incident light, while $S_{1}$, $S_{2}$ and $S_{3}$ describe the differences between polarization components along different directions. For Stokes vectors obtained under the same experimental setting, intensity normalization is commonly performed to facilitate comparison as shown in Eq(\ref{eq2}) to obtain the corresponding polarization parameters $q$, $u$ and $v$. 
\begin{equation}
\label{eq2}
\hspace{+8mm}
{q}  = \frac{S_{1}}{S_{0}}, 
{u}  = \frac{S_{2}}{S_{0}}, 
{v}  = \frac{S_{3}}{S_{0}}
\end{equation}
Based on these normalized parameters, the degree of linear polarization $DoLP$ is given by Eq(\ref{eq3}) and the angle of polarization ($AoP$) is given by Eq(\ref{eq4}).
\begin{equation}
\label{eq3}
\hspace{+14mm}
DoLP = \sqrt{q^{2} + u^{2}}
\end{equation}
The value of $DoLP$ ranges from 0 to 1 and reflects the proportion of linearly polarized light in the total intensity. In practical imaging applications, $DoLP$ is particularly sensitive to surface texture, material-dependent responses, and salient structural edges. Therefore, in this work, $S_{0}$ and $DoLP$ are selected as input modalities, so that their complementary information can be jointly exploited to generate high-quality fused images.
\begin{equation}
\label{eq4}
\hspace{+14mm}
A O P  = \frac{1}{2} \arctan \left(\frac{u}{q}\right)
\end{equation}
\subsection{Polarization image fusion methods}
\begin{figure*}
    \includegraphics[width=1\linewidth]{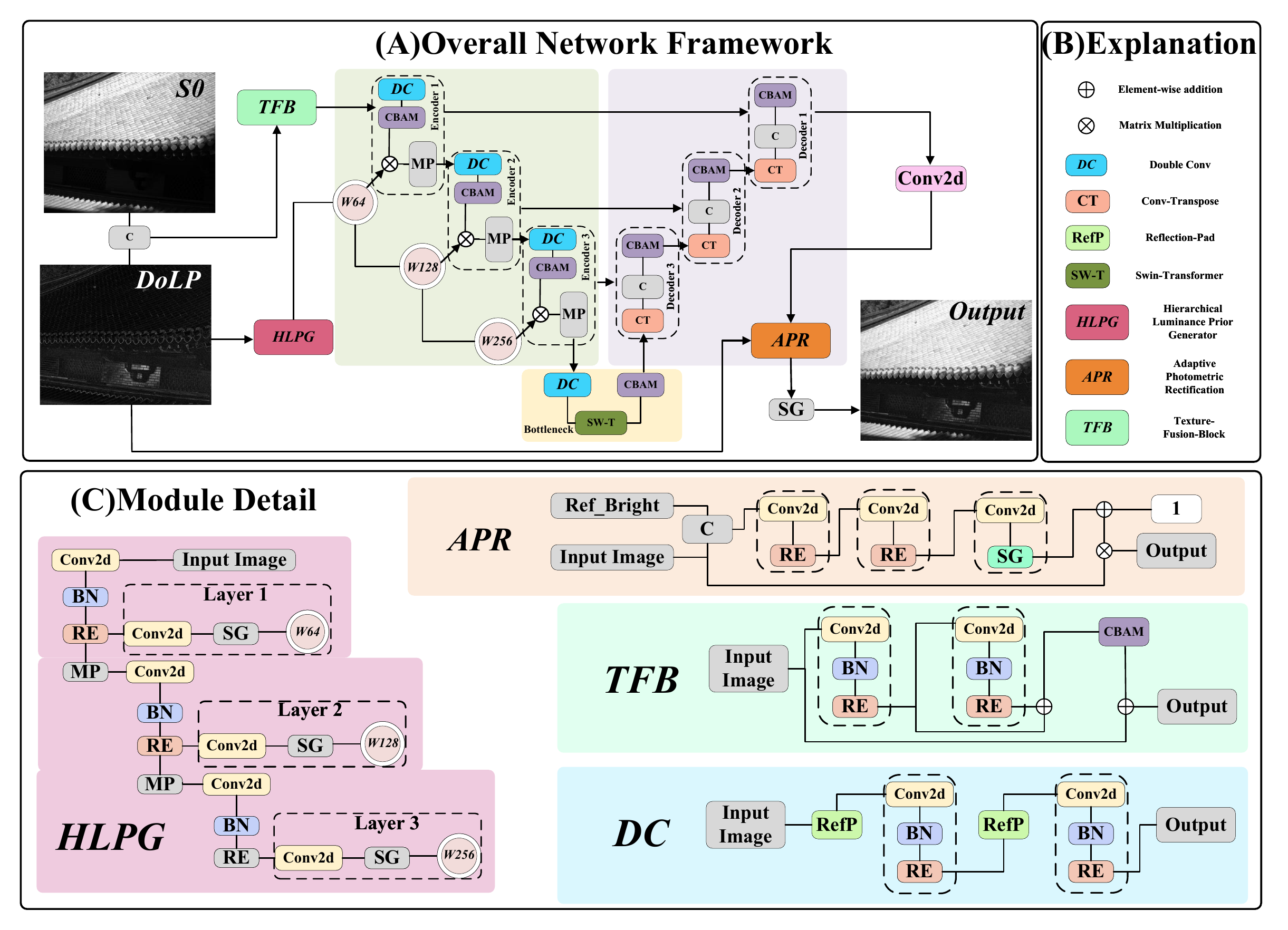}
    \vspace{-1em}
    \caption{Overview of the proposed fusion framework. The luminance-guidance module, consisting of HLPG and APR, is critical for extracting and enhancing linearly polarized details. MP: max pooling; BN: batch normalization; RE: ReLU activation; SG: Sigmoid activation.}
    \vspace{-1.5em}
    \label{fig1}
\end{figure*}

Traditional polarization image fusion methods mainly include multiscale-analysis-based methods and sparse-representation-based methods. MST-based methods exploit polarization characteristics through pyramid, wavelet, or contour-based decompositions. For example, Zhen et al. \cite{33} fused infrared intensity and polarization images using a directed Laplace pyramid, while Jiang et al. \cite{34} employed NSCT decomposition and region-based fusion rules to preserve target details. In contrast, sparse-representation-based methods improve feature extraction through adaptive dictionary learning. Li et al. \cite{35} addressed noise interference in infrared–polarization fusion by combining low-rank infrared features with sparse line-polarization features. Although these traditional methods provide certain physical interpretability, they still suffer from limited global consistency, strong dependence on hand-crafted rules, and relatively high computational cost.

Deep learning-based polarization image fusion methods can be broadly divided into CNN-based and Transformer-based frameworks. CNN-based methods were the earliest data-driven approaches in this field, mainly combining convolutional feature extraction with physical priors or task-specific constraints. Early studies focused on reflection suppression and material-detail enhancement \cite{36,37}, and later extended to more challenging conditions such as dehazing, scattering degradation, and direct polarization fusion. For example, Zhou et al. \cite{38} proposed an unsupervised polarization dehazing framework, Shi et al. \cite{39} introduced a self-supervised closed-loop optimization strategy, and Xu et al. \cite{40} developed PAPIF to better reconcile polarization and intensity information through dual attention. More recent works further incorporated hybrid and structure-aware designs, such as the CNN–Transformer fusion network of Zhang et al. \cite{41} and the frequency-decomposition framework of Liu et al. \cite{42}. Overall, CNN-based methods have substantially improved the adaptability of polarization fusion, but their reliance on local convolution still limits the modeling of long-range dependencies and global correlations among polarization features, especially under complex luminance variation.

\begin{figure*}
    \includegraphics[width=1\linewidth]{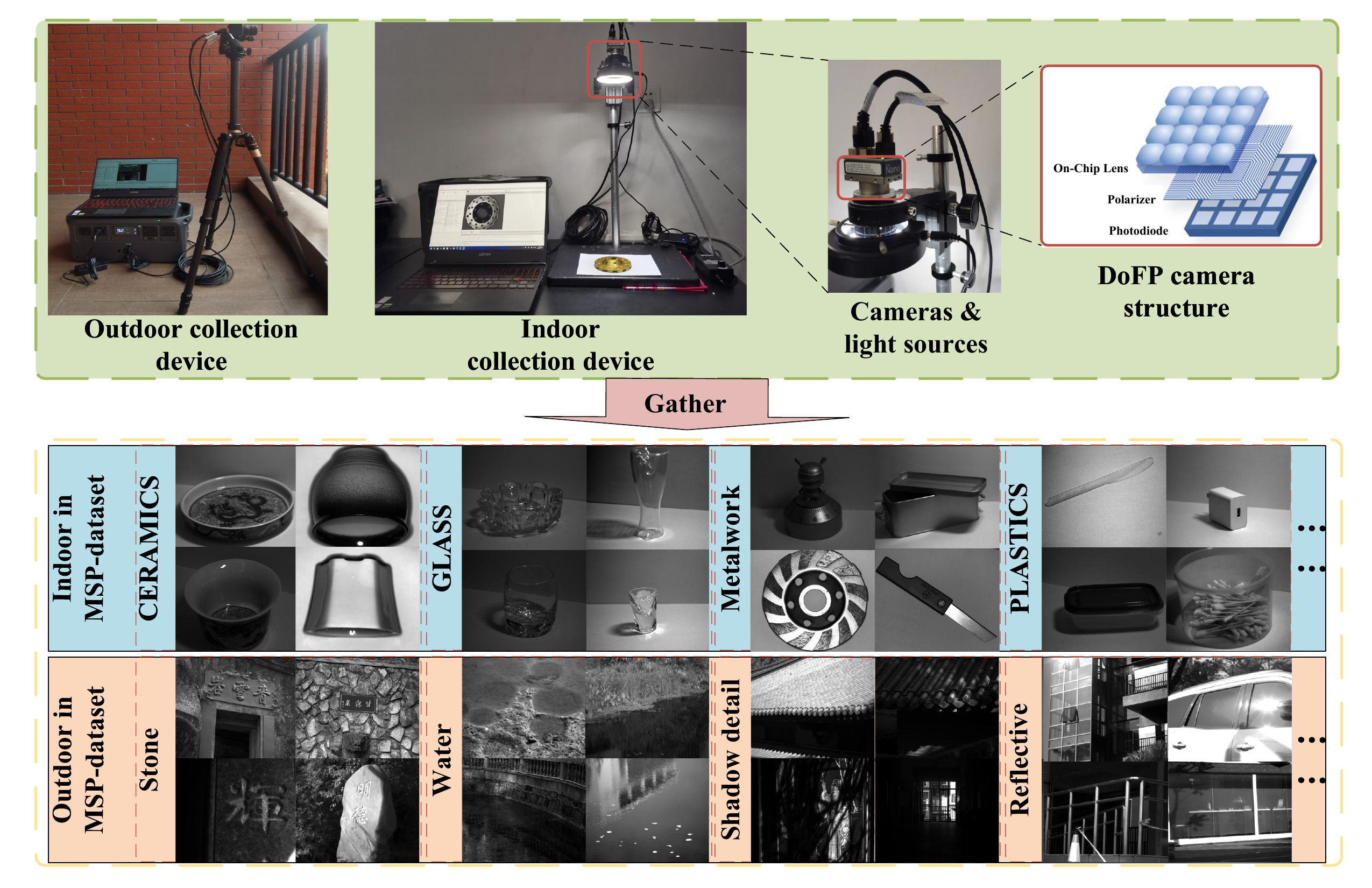}
    \vspace{-1em}
    \caption{Indoor-outdoor collection device and overview of the data set. The upper part of the figure shows the indoor and outdoor collection devices in this paper, and the lower part shows some scenes of the MSP dataset.}
    \vspace{-1.5em}
    \label{fig2}
\end{figure*}
\begin{table*}
\centering
\caption{Comparison of existing open source polarization datasets.}
\label{table1}
\begin{tabular}{cccccc}
\hline
 & \multicolumn{5}{l}{Details related to the dataset} \\ \cline{2-6} 
\multirow{-2}{*}{Dataset name} & Capturing pattern & Total scenes & Indoor Scenes & Outdoor Scenes & Image Size \\ \hline
PIF & DOFP & 74 & 5 & 69 & 1024$\times$1224 \\
GAND & DOFP & 415 & 331 & 84 & 768$\times$576 \\
MSP (Proposed) & DOFP & 1000 & 872 & 128 & 1125$\times$938 \\ \hline
\end{tabular}
\end{table*}
Transformer-based methods have been introduced into polarization imaging to better capture long-range dependencies and cross-modal interactions. Owing to the self-attention mechanism, these frameworks are more effective than purely convolutional models in modeling global contextual relationships among polarization features. Representative studies have demonstrated the potential of Transformer architectures in both polarization fusion and related polarization-aware visual tasks. Cui et al.\cite{25} proposed a SiamC-Transformer framework for shadow-robust vegetation extraction using cross-modal interaction between $DoLP$ and $S_{0}$. For direct polarization image fusion, Liu et al. \cite{49} proposed DT-F Transformer, which exploits complementary information between line-polarization images and intensity images through a dual-transpose attention mechanism. In addition, Luo et al. \cite{51} developed a lightweight Transformer-based method for color polarization image fusion, while other works further extended Transformer architectures to multimodal polarization-related tasks \cite{48,50}. Despite their improved global modeling ability, existing Transformer-based methods still show limited sensitivity to high-frequency polarization details and insufficient adaptation to complex illumination conditions.

Despite recent progress, existing methods still fail to fully exploit the key physical cues in $DoLP$ images. CNN-based methods lack sufficient global modeling ability, while Transformer-based methods remain less effective in preserving high-frequency polarization details under complex luminance conditions. To overcome these limitations, we propose a luminance-aware polarization fusion framework.
\section{Proposed method}
Given the $S_{0}$ and $DoLP$ images derived from the four-direction polarization observations $(I_{{0}^{\circ}}$, $I_{{45}^{\circ}}$, $I_{{90}^{\circ}}$, $I_{{135}^{\circ}})$ acquired by a division-of-focal-plane (DoFP) polarization camera, we develop a luminance-aware multi-scale polarization image fusion network, as illustrated in  Fig.\ref{fig1}. The proposed framework consists of three main components: TFB for multi-scale texture representation, HLPG for polarization-oriented luminance guidance, and a global-local interaction mechanism built upon CBAM and a lightweight SwinBlock. These components are integrated within a U-Net-based encoder–decoder architecture to achieve adaptive resolution recovery and cross-scale feature fusion.

The network first concatenates the input $S_{0}$ and $DoLP$ images and performs initial feature extraction using the TFB, which combines parallel convolutions, CBAM \cite{52}, and residual concatenation to capture multi-scale texture information while preserving structural details. The HLPG then extracts multi-level luminance priors from the $DoLP$ image and injects them into encoder features through interpolation and element-wise modulation. At the bottleneck, a lightweight SwinBlock is introduced to model global contextual relationships via window-based self-attention. During decoding, transposed convolution and skip connections are used to progressively recover spatial resolution and fuse multi-scale encoder features. Finally, the APR module refines the fused result using luminance-aware enhancement coefficients predicted from the input luminance information, thereby improving the balance between texture preservation and luminance consistency.

\subsection{Module}
TFB: To more effectively exploit polarization-sensitive texture information, we design TFB as the initial feature extraction unit. Given the input feature $X_0$, the first-layer feature $X_1$is obtained through convolution, batch normalization, and ReLU activation, as defined in Eq(\ref{eq5}). The same operation is then applied recursively to generate the subsequent feature representation $X_2$, using the output of the previous layer as input.

\begin{equation}
\label{eq5}
\hspace{+8mm}
X_{n}=\mathrm{ReLU} (BN+(\mathrm{Conv} (X_{n-1})))
\end{equation}

After hierarchical feature extraction, intermediate features are aggregated and refined by CBAM to enhance informative responses. The refined feature is then combined with the input feature through a residual connection, and the final output $X_{final}$is obtained by ReLU activation, as formulated in Eq(\ref{eq6}). This design enables the TFB to preserve original structural information while enhancing texture-sensitive responses in polarization images.

\begin{equation}
X_{final}=\mathrm{ReLU} (X_{0}+(\mathrm{CBAM} (X_{n}+X_{n-1})))
\label{eq6}
\hspace{+8mm}
\end{equation}

In addition, to improve sensitivity to boundary and edge details in the generated image, the encoder adopts the double-convolution module commonly used in image segmentation networks as the basic feature extraction unit of the U-Net architecture. Specifically, reflection padding is used to alleviate boundary artifacts, while convolution, batch normalization, and nonlinear activation are combined to enhance feature representation while preserving boundary information.

Luminance guidance section: Because $DoLP$ image carries polarization-sensitive cues related to surface texture and material-dependent appearance, it provides an important source of complementary information in polarization image fusion. To better exploit this property, we design two luminance-guidance modules for linearly polarized information modeling. The first one is the APR module, which is inspired by the use of reference luminance maps in Enlighten-GAN \cite{53} and HDR-GAN \cite{54} for enhancement guidance. Different from existing luminance-enhancement strategies, the proposed APR introduces luminance normalization based on the texture characteristics of $DoLP$ images and adopts a lightweight convolutional structure to improve robustness under varying illumination conditions. The detailed formulation is given below.
Given a reference luminance map $B_{ref}$, defined as the channel-wise mean of the $DoLP$ image, normalization is first performed to obtain $B_{nor}$, as formulated in Eq(\ref{eq7}), where $\epsilon=1×10^{-6}$ is used for numerical stability.

\begin{equation}
B_{nor}=\frac{B_{ref}-\min \left(B_{ref}\right)}{\max \left(B_{ref}\right)-\min \left(B_{ref}\right)+\epsilon} \in[0,1]
\label{eq7}
\end{equation}

The fused feature map $X\epsilon R^{C×H×w}$ is then concatenated with $B_{nor}$ along the channel dimension to obtain an augmented feature representation $X\epsilon R^{{(C+1)}×H×w}$. Based on this representation, the luminance attention coefficient$M$ is predicted according to Eq(\ref{eq8}), where $W(X)$ denotes a convolution followed by ReLU activation $\sigma$ denotes the Sigmoid function.
\begin{equation}
\hspace{+12mm}
M=\mathrm{\sigma } \{\mathrm{Conv}[ W(W(X))] \}
\label{eq8}
\end{equation}

Finally, $X_{enhanced}$ is output by Eq(\ref{eq9}) to realize the luminance adaptive correction to the feature map, where $\oplus $ denotes element-wise multiplication.
\begin{equation}
\hspace{+13mm}
X_{enhanced}=\mathrm{X} \oplus \mathrm{(1+M)}
\label{eq9}
\end{equation}

The second luminance-guidance module is the HLPG, which is designed to generate multi-scale luminance priors from $DoLP$ images. Its design is inspired by multi-exposure fusion strategies\cite{55}, where different luminance distributions are integrated through scale-dependent weighting. In contrast to applying luminance guidance only at the final reconstruction stage, HLPG injects hierarchical luminance priors into multiple stages of the encoder, thereby providing luminance-aware modulation throughout feature extraction. Following the three-stage downsampling structure of the encoder, HLPG employs a three-layer convolutional network to extract luminance features at different scales and to generate the corresponding spatial attention weights, as formulated in Eqs.\ref{eq10}–\ref{eq11}. Together with APR, HLPG forms a complete luminance-guidance chain that enables the network to more effectively exploit brightness-related information in $DoLP$ images under complex illumination.

\begin{equation}
f_{n}=\left\{\begin{array}{c}
\mathrm{ReLU}\{B N[\mathrm{Conv}(X)]\}, n=1 \\
\mathrm{ReLU}\left\{B N\left[\mathrm{Conv}\left(\mathrm{Pool}\left(f_{1}\right)\right)\right]\right\}, n=2 \\
\mathrm{ReLU}\left\{B N\left[\mathrm{Conv}\left(\mathrm{Pool}\left(f_{2}\right)\right)\right]\right\}, n=3
\end{array}\right.
\label{eq10}
\end{equation}
\begin{equation}
\hspace{+6mm}
\omega _{n}=\left\{\begin{array}{c}
\mathrm{\sigma }\left[{Conv}\left(f_{1}\right)\right], n=64 \\
\mathrm{\sigma }\left[{Conv}\left(f_{2}\right)\right], n=128 \\
\mathrm{\sigma }\left[{Conv}\left(f_{3}\right)\right], n=256 \\
\end{array}\right.
\label{eq11}
\end{equation}
In addition, the proposed network employs CBAM and a lightweight SwinBlock for attention-based feature refinement. CBAM enhances informative features through cascaded channel and spatial attention, while the lightweight SwinBlock captures broader contextual relationships with reduced computational complexity. By removing positional encoding and the window-shift mechanism, the lightweight design improves efficiency while retaining nonlocal modeling ability. The combination of the two modules provides a better trade-off between feature sensitivity and computational cost.
\subsection{Loss function}

To better preserve dominant texture details and balance the contributions of different polarization characteristics during fusion, we design a multi-objective joint optimization loss function, denoted by $L_{all}$. This loss jointly constrains structural similarity, pixel-level fidelity, local contrast, texture preservation, and model complexity, and is defined as follows:
\begin{equation}
\hspace{-1mm}
{L}_{\text {all }}=\lambda_{1} L_{\text {SSIM }}+\lambda_{2} L_{L 1}
+\lambda_{3} L_{\text {CON }}+\lambda_{4} L_{\text {TEX }}+\lambda_{5} L_{\text {Reg }} 
\label{eq12}
\end{equation}
where $\lambda_{1\sim 5} $ denote the weighting coefficients of the five loss terms, which are set to 0.6, 0.15, 0.1, 0.1, and 0.05, respectively.

\begin{figure*}
    \includegraphics[width=1\linewidth]{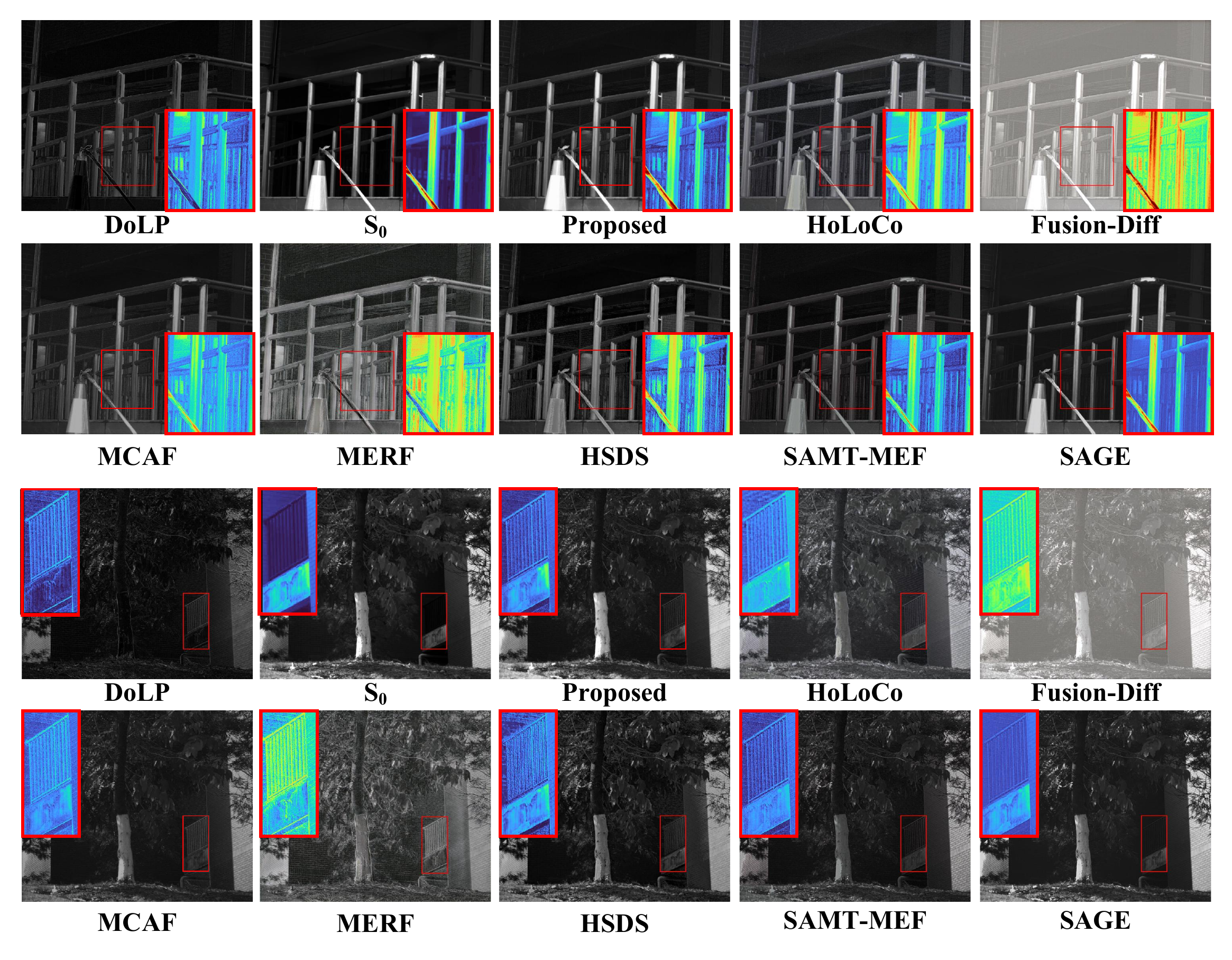}
    \vspace{-2em}
    \caption{Example demonstration of ironwork detail on the MSP dataset.}
    \vspace{-1.5em}
    \label{fig3}
\end{figure*}

$L_{{SSIM}}$ measures the structural similarity between the fused image and the source images, thereby encouraging the preservation of structural details during fusion. It is defined as follows:
\begin{equation}
\mathcal{L}_{\text {SSIM }}=\frac{1}{2} \sum_{k=1}^{2}\left[1-\text { SSIM }\left(I_{\text {pred }}, I_{\text {target }}^{k}\right)\right]
\label{eq13}
\end{equation}
where $I_{pred}$ denotes the fused output image, and $I_{\text {target }}^{k}$ denotes
the input source image, k=0 or k=1 corresponding $S_{0}$ and $DoLP$ images, respectively. The structural similarity ${SSIM}(x, y)$ is computed as shown in Eq(\ref{eq14}), where $\mu_{x}$,$\mu_{y}$ denote the local mean of the images x and y, $\sigma_{x}^{2}$,$\sigma_{y}^{2}$ denote the corresponding variances, $\sigma_{x y}$ denotes the covariance, and  $C_{1}$  and $C_{2}$ are custom constants used for stabilization calculation.
\begin{equation}
\hspace{-2mm}
\operatorname{SSIM}(x, y)=\frac{\left(2 \mu_{x} \mu_{y}+C_{1}\right)\left(2 \sigma_{x y}+C_{2}\right)}{\left(\mu_{x}^{2}+\mu_{y}^{2}+C_{1}\right)\left(\sigma_{x}^{2}+\sigma_{y}^{2}+C_{2}\right)}
\label{eq14}
\end{equation}

$L_{L 1}$ denotes the pixel-level absolute error loss. It is introduced to constrain the global luminance consistency of the fused image with respect to the source images, while also suppressing dark-noise amplification in the $DoLP$-guided fusion process. It is defined as follows:
\begin{equation}
\hspace{+10mm}
\mathcal{L}_{\text {L 1 }}=\frac{1}{2} \sum_{k=1}^{2} \left |  I_{\text {pred }}- I_{\text {target }}^{k} \right |
\label{eq15}
\end{equation}

${L}_{\text {CON }}$ is introduced to preserve local contrast and prevent polarization-sensitive details from becoming over-smoothed during fusion. By directly constraining luminance variation, this loss enhances the distinction between bright and dark regions in the fused image. It is defined as follows:
\begin{small}
\begin{equation}
\hspace{-2mm}
\mathcal{L}_{\text {CON }}=\max \left(0,1-\sqrt{\frac{1}{H * W} \sum_{i=1}^{H} \sum_{j=1}^{W}\left(X_{n, c, i, j}-\mu_{n, c}\right)^{2}+\varepsilon}\right)
\label{eq16}
\end{equation}
\end{small}where $\mu_{n, c}=\frac{1}{H * W}\sum_{i,j}X_{n, c, i, j}$  denotes the channel-wise mean of the input image, $\varepsilon $ is a very small positive number used to stabilize the values,  $ X_{n, c, i, j} $ denotes the tensor element at location $( i, j)$ in channel $c$. 

In addition, $L_{TEX}$ is used as a texture-preserving constraint by comparing the horizontal and vertical gradient maps of the fused image and the target image. This loss encourages the network to generate sharper fused results by maintaining gradient consistency in both directions. It is defined as follows:
\begin{small}
\begin{equation}
\hspace{-2mm}
\mathcal{L}_{T E X}=\frac{1}{2}\left(\left\|\nabla_{x} I_{\text {pred }}-\nabla_{x} I_{\text {target }}\right\|_{1}+\left\|\nabla_{y} I_{\text {pred }}-\nabla_{y} I_{\text {target }}\right\|_{1}\right)
\label{eq17}
\end{equation}
\end{small}

Finally, $L_{Reg}$ is introduced to control model complexity and reduce the risk of overfitting, thereby improving generalization ability. It is defined as follows:
\begin{equation}
\hspace{+16mm}
\label{eq18}
\mathcal{L}_{\text {Reg }}=\sum_{t=1}^{T}\left\|\theta_{t}\right\|_{2}
\end{equation}

This regularization term is implemented by computing the $L_{2}$ norm of the learnable model parameters, where $\left \| X \right \|_{2}$ denotes the Euclidean norm and $\theta_{t}$ denotes the t layer learnable parameter in the model.
\begin{figure*}
\centering
\includegraphics[width=1.0\linewidth]{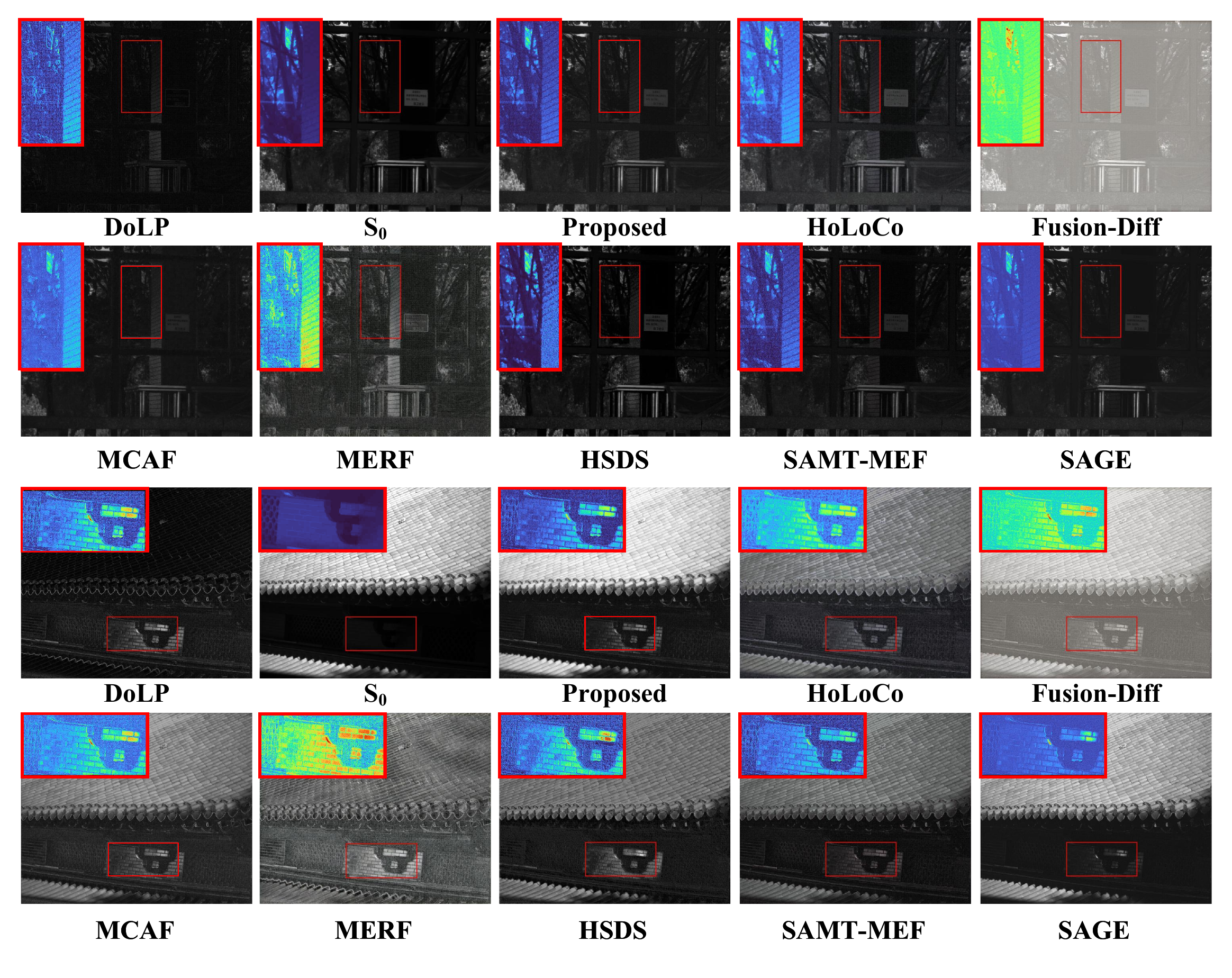}
\vspace{-2em}
\caption{Example demonstration of dark detail on the MSP dataset.}
\vspace{-1.5em}
\label{fig4}
\end{figure*}
\begin{table*}[htbp]
\centering
{\rmfamily 
\caption{Mean values of the metrics on the MSP dataset for the different fusion methods. (Red: optimal, blue: second best, green: third best).}
\label{table2}
\begin{tabular}{ccccccc}
\toprule
 & \multicolumn{6}{l}{Metrics} \\ \cmidrule(lr){2-7} 
\multirow{-2}{*}{Methods} & $SSIM$ & $VIF$ & $SD$ & $MS$-$SSIM$ & $Q_{MI}$ & $Q^{ab/f}$ \\ 
\midrule
HoLoCo(2023) & 0.513 & 0.224 & 30.753 & 0.876 & 0.308 & 0.274 \\
Fusion-Diff(2023) & 0.404 & 0.256 & 30.129 & {\color[HTML]{34FF34} 0.905} & 0.330 & 0.298 \\
MCAF (2023) & 0.609 & {\color[HTML]{34CDF9} 0.367} & 35.305 & 0.906 & {\color[HTML]{34FF34} 0.438} & {\color[HTML]{34CDF9} 0.478} \\
MERF(2024) & 0.357 & 0.216 & 32.573 & 0.626 & 0.235 & 0.353 \\
HSDS(2024) & 0.538 & 0.214 & {\color[HTML]{34FF34} 37.348} & 0.858 & 0.349 & 0.346 \\
SAMT-MEF(2024) & {\color[HTML]{34CDF9} 0.615} & 0.333 & 31.531 & {\color[HTML]{34CDF9} 0.929} & 0.378 & {\color[HTML]{FE0000} 0.484} \\
SAGE(2025) & {\color[HTML]{34FF34} 0.611} & {\color[HTML]{34FF34} 0.356} & {\color[HTML]{34CDF9} 45.526} & 0.894 & {\color[HTML]{FE0000} 0.496} & 0.366 \\
Proposed & {\color[HTML]{FE0000} 0.637} & {\color[HTML]{FE0000} 0.440} & {\color[HTML]{FE0000} 55.802} & {\color[HTML]{FE0000} 0.930} & {\color[HTML]{34CDF9} 0.491} & {\color[HTML]{34FF34} 0.454} \\
\bottomrule
\end{tabular}
}
\end{table*}

\section{Experiments}
\subsection{Experimental settings}
\subsubsection{Dataset construction}
To evaluate the proposed method under complex real-world lighting conditions, we construct a new multi-scene polarization image fusion dataset, termed MSP. As summarized in Table \ref{table1}, MSP contains 1,000 image pairs captured using a division-of-focal-plane (DoFP) polarization imaging system, covering 17 types of indoor and outdoor illumination scenarios. Compared with existing public datasets such as PIF \cite{49} and GAND \cite{47}, MSP provides broader coverage in terms of scene diversity, material categories, and lighting complexity, including 872 indoor scenes and 128 outdoor scenes. Representative examples of the acquisition setup and dataset content are shown in Fig.\ref{fig2}.

For each captured raw image with a resolution of 2464 × 2056, we first separate the four polarization-direction sub-images according to the orientations of the embedded micro-polarizers. The spatial resolution is then recovered by bilinear interpolation, after which the corresponding polarization components are computed according to the Stokes-parameter formulation introduced in Section 2. As a result, each sample in MSP contains four-direction polarization observations $(I_{{0}^{\circ}}$, $I_{{45}^{\circ}},$ $I_{{90}^{\circ}}$, $I_{{135}^{\circ}})$, together with the derived $S_{0}$,  $DoLP$ , and $AoP$ images.

To enhance the practical relevance of the dataset, MSP is mainly collected from materials that exhibit pronounced polarization responses, including plastics, metal products, stones, glass, glazes, ceramics, and sand. In addition, the dataset covers diverse illumination conditions such as daytime and nighttime scenes, strong natural illumination, and exposure variations ranging from overexposed to normally exposed and underexposed conditions. The collected data also include a variety of realistic outdoor environments, such as seashores, sandy beaches, woodlands, temples, and roads. These characteristics make MSP a challenging benchmark for evaluating polarization image fusion in material-sensitive and luminance-varying scenarios.

\subsubsection{Training details}

For training and evaluation on the MSP dataset, 450 samples were randomly selected from each source-modality pair
($S_{0}$, $DoLP$). Among them, 400 samples were used for training and 50 samples for validation, while the remaining 600 samples in MSP were used as the test set. Since the proposed method is intended to enhance polarization-sensitive material details under complex real-world lighting conditions, indoor and outdoor scenes are not explicitly separated during training; instead, all scenes are uniformly used to train and evaluate the model.

During training, the batch size was set to 4 and the number of training epochs was set to 335. Adam was adopted as the optimizer with an initial learning rate of 0.0001. All experiments were conducted on a workstation equipped with an NVIDIA GeForce RTX 3090 GPU.

\subsection{Comparative experiments}
\subsubsection{Compared methods}

To comprehensively evaluate the effectiveness of the proposed method, we compare it with seven representative fusion methods, including HoLoCo \cite{56}, Fusion-Diff \cite{57}, MERF \cite{58}, HSDS \cite{59}, SAGE \cite{60}, SAMT-MEF \cite{61} and MCAF \cite{62}.As shown in Table \ref{table2}, these baselines cover a range of recent image fusion paradigms and provide diverse reference points for evaluating performance under different scene conditions.

Among them, HoLoCo and Fusion-Diff represent two recent fusion architectures with different modeling strategies, while MCAF and MERF further emphasize cross-scale information interaction and progressive feature refinement. HSDS and SAMT-MEF provide additional strong baselines for evaluating robustness and structural preservation in complex scenes, and SAGE serves as the most recent high-performing reference among the compared methods. By comparing against these representative approaches, the effectiveness of the proposed framework can be evaluated more objectively from the perspectives of detail preservation, luminance consistency, and overall fusion quality.

\subsubsection{Performance metrics}
In this paper, six objective metrics are adopted in the fusion performance from both the physical-property representation and visual-perception perspectives: SSIM (Structural Similarity) \cite{63}, VIF (Visual Information Fidelity) \cite{64}, SD (Standard Deviation) \cite{65}, MS-SSIM (Multiscale Structural Similarity) \cite{66}, $Q^{ab/f}$ (Fusion Quality) and $Q_{MI}$ (Normalized Mutual Information) \cite{67}.These metrics jointly assess structural consistency, visual information fidelity, contrast richness, and information preservation in the fused results. 
\begin{figure*}
\includegraphics[width=1\linewidth]{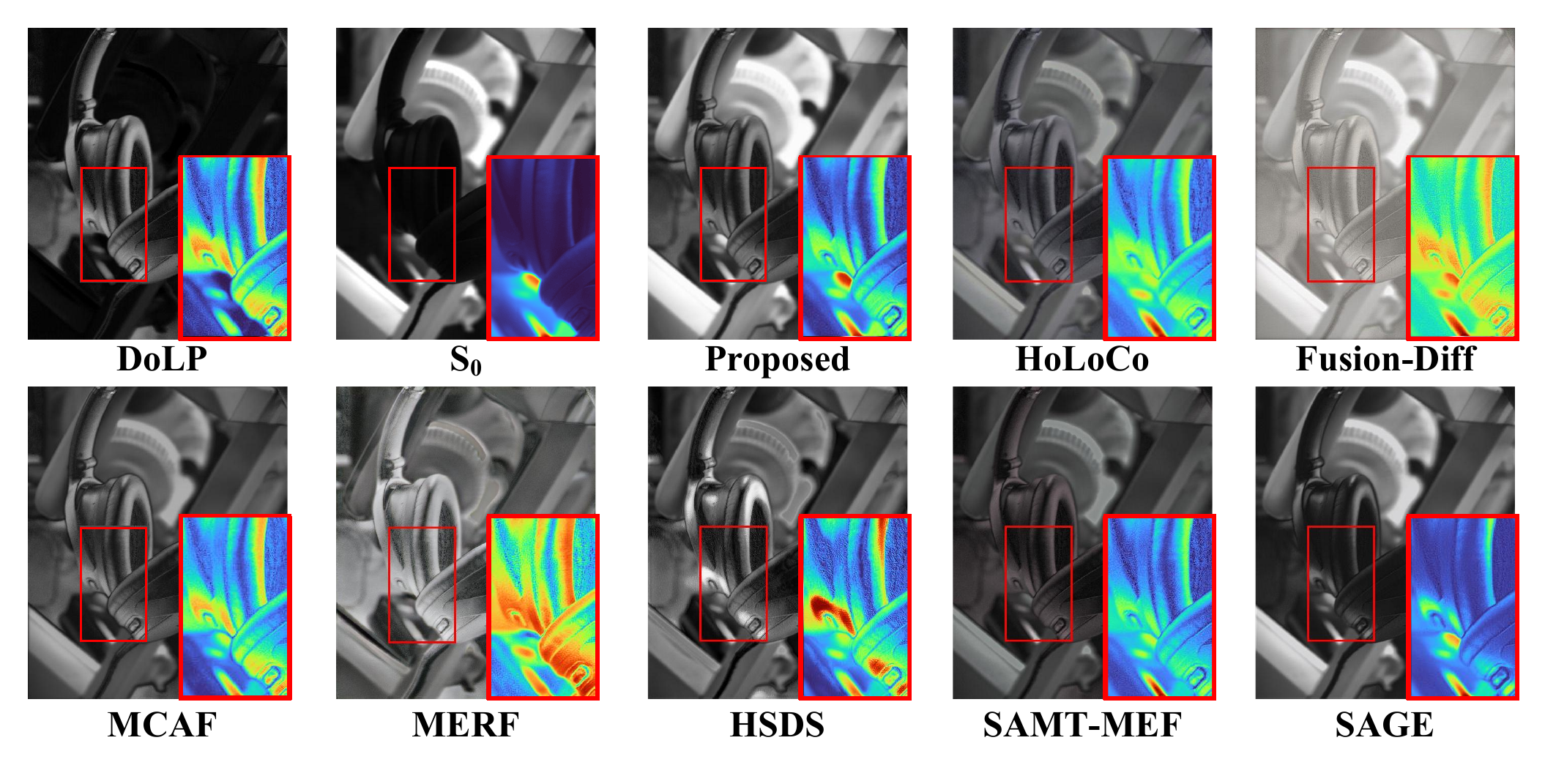}
\vspace{-2em}
\caption{Example of a holster material on a PIF dataset.}
\vspace{-1.5em}
\label{fig5}
\end{figure*}

\begin{figure*}
\includegraphics[width=1\linewidth]{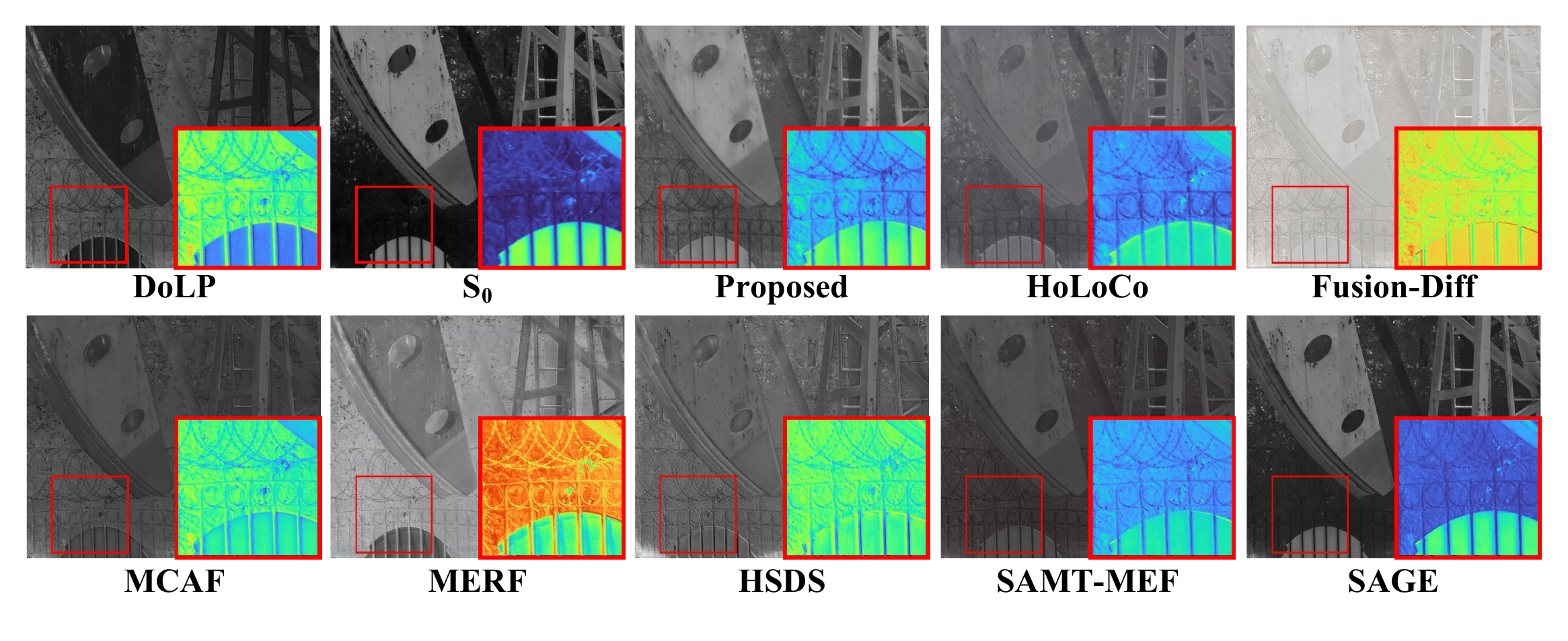}
\vspace{-2em}
\caption{Example demonstration of ironwork details on the GAND dataset.}
\vspace{-1.5em}
\label{fig6}
\end{figure*}
\begin{table*}[htbp]
\centering
{\rmfamily
\caption{Mean values of the metrics on the PIF dataset for the different fusion methods.}
\label{table3}
\begin{tabular}{ccccccc}
\toprule
 & \multicolumn{6}{l}{Metrics} \\ \cmidrule(lr){2-7} 
\multirow{-2}{*}{Methods} & $SSIM$ & $VIF$ & $SD$ & $MS$-$SSIM$ & $Q_{MI}$ & $Q^{ab/f}$ \\ 
\midrule
HoLoCo(2023)   & 0.555 & 0.327 & 28.400 & 0.890 & 0.358 & 0.378 \\
Fusion-Diff(2023) & 0.411 & 0.371 & 29.258 & {\color[HTML]{34CDF9} 0.911} & 0.382 & {\color[HTML]{34FF34} 0.491} \\
MCAF(2023)     & 0.609 & 0.341 & 32.974 & 0.889 & {\color[HTML]{34FF34} 0.468} & 0.377 \\
MERF(2024)     & 0.370 & 0.217 & 33.028 & 0.599 & 0.260 & 0.381 \\
HSDS(2024)     & 0.540 & 0.224 & {\color[HTML]{34FF34} 38.154} & 0.876 & 0.327 & 0.416 \\
SAMT-MEF(2024) & {\color[HTML]{FE0000} 0.647} & {\color[HTML]{34FF34} 0.407} & 28.042 & {\color[HTML]{34FF34} 0.911} & 0.434 & {\color[HTML]{34CDF9} 0.498} \\
SAGE(2025)     & {\color[HTML]{34CDF9} 0.640} & {\color[HTML]{FE0000} 0.470} & {\color[HTML]{34CDF9} 43.325} & 0.902 & {\color[HTML]{FE0000} 0.542} & 0.463 \\
Proposed       & {\color[HTML]{34FF34} 0.623} & {\color[HTML]{34CDF9} 0.413} & {\color[HTML]{FE0000} 54.474} & {\color[HTML]{FE0000} 0.942} & {\color[HTML]{34CDF9} 0.504} & {\color[HTML]{FE0000} 0.521} \\
\bottomrule
\end{tabular}
} 
\end{table*}

\subsubsection{Qualitative comparison of fusion results}

To qualitatively evaluate the fusion performance, four representative scenes from three datasets were selected for visual comparison, including two samples from MSP, one sample from PIF, and one sample from GAND. Since both $S_{0}$ and $DoLP$ are grayscale images, the zoomed-in regions are additionally visualized using the HSV colormap in MATLAB to improve the perceptual visibility of subtle structural differences. It should be emphasized that this pseudo-color rendering is used only for visual presentation and does not affect network training, inference, or quantitative evaluation.
\begin{table*}[htbp]
\centering
{\rmfamily
\caption{Mean values of the metrics on the GAND dataset for the different fusion methods. }
\label{table4}
\begin{tabular}{ccccccc}
\toprule
 & \multicolumn{6}{l}{Metrics} \\ \cmidrule(lr){2-7} 
\multirow{-2}{*}{Methods} & $SSIM$ & $VIF$ & $SD$ & $MS$-$SSIM$ & $Q_{MI}$ & $Q^{ab/f}$ \\ 
\midrule
HoLoCo(2023)   & 0.565 & 0.251 & 18.254 & 0.785 & 0.226 & 0.215 \\
Fusion-Diff(2023) & 0.450 & {\color[HTML]{34FF34} 0.295} & 19.449 & 0.802 & 0.253 & 0.325 \\
MCAF(2023)     & {\color[HTML]{34CDF9} 0.616} & 0.290 & 21.934 & 0.708 & {\color[HTML]{34FF34} 0.324} & 0.318 \\
MERF(2024)     & 0.410 & 0.247 & {\color[HTML]{34FF34} 23.685} & 0.359 & 0.270 & {\color[HTML]{34CDF9} 0.365} \\
HSDS(2024)     & 0.538 & 0.164 & 23.152 & 0.751 & 0.167 & 0.293 \\
SAMT-MEF(2024) & {\color[HTML]{FE0000} 0.634} & 0.277 & 19.555 & {\color[HTML]{34FF34} 0.819} & 0.278 & 0.264 \\
SAGE(2025)     & {\color[HTML]{34FF34} 0.617} & {\color[HTML]{34CDF9} 0.300} & {\color[HTML]{FE0000} 37.260} & {\color[HTML]{34CDF9} 0.879} & {\color[HTML]{FE0000} 0.427} & {\color[HTML]{FE0000} 0.424} \\
Proposed       & 0.613 & {\color[HTML]{FE0000} 0.301} & {\color[HTML]{34CDF9} 35.997} & {\color[HTML]{FE0000} 0.891} & {\color[HTML]{34CDF9} 0.343} & {\color[HTML]{34FF34} 0.360} \\
\bottomrule
\end{tabular}
}
\end{table*}

The two scenes shown in Fig.\ref{fig3} mainly reflect the sensitivity of polarization imaging to special materials in complex environments. In the first scene, HoLoCo, Fusion-Diff, MERF, and SAMT-MEF exhibit different degrees of detail loss in the near-field dark metal railings, while MCAF and HSDS preserve part of the railing texture but still show noticeable color deviation from the $S_{0}$image. By contrast, the proposed method better preserves the linearly polarized details of the iron railings while retaining the ambient luminance information of the intensity image. More importantly, in the second scene, the proposed method more effectively enhances the texture details of the distant metal railing while simultaneously preserving the shadow details of the illuminated wall on the lower left side and the structural details of the wall on the right side. These results indicate that the proposed method is more effective in preserving material-sensitive details in linearly polarized images under complex luminance conditions.

Figure \ref{fig4} mainly demonstrates the recovery of dark-region details. In the first scene, SAGE and HSDS show substantial loss of the tile patterns in the reflected region, while HoLoCo, Fusion-Diff, and MERF exhibit obvious luminance distortion, making the black contour of the white plate on the column less distinguishable. Although MCAF and SAMT-MEF preserve part of the structural information, they fail to effectively enhance the material-sensitive details of the iron table below the column. By contrast, the proposed method better preserves both reflected structures and material-related texture cues. In the second scene, SAMT-MEF shows obvious color distortion, while MERF, HoLoCo, and Fusion-Diff produce excessively high overall brightness. MCAF and HSDS also suffer from severe noise interference in the grid-like reflective wall region, which degrades fine textures. By comparison, the proposed method better preserves the detail of the eave end at the junction between the backlit and illuminated regions while showing stronger suppression of noise interference, indicating more effective preservation of discriminative details in low-light environments.

To further assess generalization ability, we additionally present representative results on PIF and GAND. In the PIF example shown in Fig.\ref{fig5}, the proposed method better preserves the fine texture of key material regions, such as the headphone sponge cushion and the plastic shell, while also maintaining the white fan contour in the background of the $S_{0}$ image. By contrast, several competing methods show varying degrees of structural attenuation or visual distortion. In the GAND example shown in Fig.\ref{fig6}, which contains complex outdoor iron railing details, HoLoCo, Fusion-Diff, MERF, and SAMT-MEF exhibit different degrees of visual distortion, whereas MCAF and HSDS show stronger noise interference in the railing region and noticeable degradation around the semicircular plastic baffle in the middle of the scene. Overall, the proposed method achieves a more favorable balance between detail enhancement, structural fidelity, and luminance consistency across the three datasets.

\begin{table*}[htb]
\centering
{\rmfamily
\caption{Test the mean metrics of different module models in the MSP dataset.}
\label{table5}
\begin{tabular}{ccccccc}
\toprule
 & \multicolumn{6}{l}{Metrics} \\ \cmidrule(lr){2-7}
\multirow{-2}{*}{Methods} & $SSIM$ & $VIF$ & $SD$ & $MS$-$SSIM$ & $Q_{MI}$ & $Q^{ab/f}$ \\ 
\midrule
Total                   & 0.637 & 0.440 & {\color[HTML]{34CDF9} 55.802} & {\color[HTML]{FE0000} 0.930} & {\color[HTML]{FE0000} 0.491} & {\color[HTML]{FE0000} 0.454} \\
base+CBAM+LUG         & 0.655 & 0.447 & {\color[HTML]{FE0000} 57.170} & {\color[HTML]{34CDF9} 0.926} & {\color[HTML]{34CDF9} 0.487} & {\color[HTML]{34CDF9} 0.449} \\
base+TFB+LUG         & 0.659 & {\color[HTML]{34CDF9} 0.476} & 49.169 & 0.919 & {\color[HTML]{34FF34} 0.479} & 0.433 \\
base+CBAM+TFB           & 0.648 & 0.461 & 50.783 & 0.912 & 0.458 & 0.408 \\
base+CBAM                & {\color[HTML]{34CDF9} 0.662} & 0.460 & {\color[HTML]{34FF34} 52.317} & {\color[HTML]{34FF34} 0.920} & 0.462 & 0.443 \\
base+TFB                & 0.656 & 0.469 & 52.048 & 0.918 & 0.500 & {\color[HTML]{34FF34} 0.448} \\
base+LUG              & {\color[HTML]{FE0000} 0.663} & {\color[HTML]{34FF34} 0.470} & 50.010 & 0.906 & 0.459 & 0.430 \\
base                     & {\color[HTML]{34FF34} 0.661} & {\color[HTML]{FE0000} 0.484} & 49.132 & 0.916 & 0.475 & 0.420 \\ 
\bottomrule
\end{tabular}
}
\end{table*}

\begin{table*}[htbp]
\centering
{\rmfamily
\caption{Testing the impact of different components of the loss function on the MSP dataset.}
\label{table6}
\begin{tabular}{ccccccc}
\toprule
 & \multicolumn{6}{l}{Metrics} \\ \cmidrule(lr){2-7}
\multirow{-2}{*}{Methods} & $SSIM$ & $VIF$ & $SD$ & $MS$-$SSIM$ & $Q_{MI}$ & $Q^{ab/f}$ \\ 
\midrule
No-Reg                                            & {\color[HTML]{34FF34} 0.648} & {\color[HTML]{FE0000} 0.470} & {\color[HTML]{34FF34} 50.515} & {\color[HTML]{34CDF9} 0.917} & {\color[HTML]{34FF34} 0.471} & 0.406                        \\
No-L1                                              & {\color[HTML]{34CDF9} 0.649} & {\color[HTML]{34CDF9} 0.469} & {\color[HTML]{34CDF9} 53.251} & 0.912                        & 0.459                        & {\color[HTML]{34FF34} 0.414} \\
No-TEX                                             & {\color[HTML]{FE0000} 0.657} & {\color[HTML]{34FF34} 0.468} & 47.354                        & {\color[HTML]{34FF34} 0.916} & {\color[HTML]{34CDF9} 0.488} & {\color[HTML]{34CDF9} 0.417} \\
Total                                             & 0.637                        & 0.440                        & {\color[HTML]{FE0000} 55.802} & {\color[HTML]{FE0000} 0.930} & {\color[HTML]{FE0000} 0.491} & {\color[HTML]{FE0000} 0.454} \\ \bottomrule
\end{tabular}
}
\end{table*}

\begin{table*}[htbp]
\centering
{\rmfamily
\caption{Comparison of computational efficiency.}
\label{table7}
\begin{tabular}{ccccccc}
\toprule
 & \multicolumn{3}{l}{Computational efficiency} \\ \cmidrule(lr){2-4}
\multirow{-2}{*}{Methods} & Time (s) & FLOPs (G) & Parameters (M) \\ 
\midrule
HoLoCo (2023)        & 3.360  & {\color[HTML]{34CDF9} 100.377} & {\color[HTML]{FE0000} 0.114} \\
Fusion-Diff (2023)   & 65.590 & {\color[HTML]{34FF34} 239.857} & 26.899 \\
MCAF (2023)          & 6.080  & 368.001 & {\color[HTML]{34FF34} 0.233} \\
HSDS (2024)          & 19.330 & 925.011 & 1.166 \\
SAMT-MEF (2024)      & {\color[HTML]{34CDF9} 0.163} & 999.200 & 1.230 \\
SAGE (2025)          & {\color[HTML]{FE0000} 0.020} & {\color[HTML]{FE0000} 69.838} & {\color[HTML]{34CDF9} 0.136} \\
Proposed             & {\color[HTML]{34FF34} 1.610} & 791.314 & 11.120 \\ 
\bottomrule
\end{tabular}
}
\end{table*}

In summary, the proposed method shows clear advantages in both qualitative and quantitative evaluations. As reported in Table \ref{table2}, for the 599 image pairs selected from the MSP dataset, the proposed method achieves the best results in SSIM, MS-SSIM, VIF, and SD. In particular, the clear improvements in SD and VIF indicate richer detail representation and stronger consistency with the source-image information. Although $Q^{ab/f}$ and $Q_{MI}$ do not reach the best values, they still remain among the top-performing results, which further supports the effectiveness of the proposed method.

As shown in Table \ref{table3}, for the 40 image pairs selected from the PIF dataset, the proposed method achieves the best performance in MS-SSIM, $Q^{ab/f}$, and SD, while VIF and $Q_{MI}$ rank second among the compared methods. These results indicate that the proposed method better preserves salient source information and structural details, while also demonstrating strong generalizability on the PIF dataset. Table \ref{table4} reports the average metric values for 40 image pairs selected from the GAND dataset. On this dataset, the proposed method achieves the best results in MS-SSIM and VIF, while SD and $Q_{MI}$ rank second. This suggests that the proposed method remains highly competitive on GAND and maintains stable performance in terms of structural preservation and visual information fidelity.

In general, the proposed method achieves consistently competitive results in all three datasets and shows the clearest advantage on MSP. These findings demonstrate its effectiveness and robustness for polarization image fusion under complex luminance conditions.

\subsection{Ablation experiments}

To verify the effectiveness of the proposed framework, ablation experiments are conducted on MSP from two perspectives: module design and loss-function design. For module ablation, three key components are considered, namely CBAM, TFB, and the luminance-guidance module (LUG) formed by HLPG and APR. By enabling or removing these components, eight model variants are constructed, and the quantitative results are reported in Table \ref{table5}.

As shown in Table \ref{table5}, the complete model achieves the best performance in MS-SSIM , $Q_{MI}$ and $Q^{ab/f}$, while remaining competitive in SD . Although some reduced variants perform slightly better in individual metrics such as SSIM or VIF, the complete model provides a more balanced overall performance. In particular, configurations containing the luminance-guidance module generally show stronger results in SD and MS-SSIM, demonstrating the importance of luminance-aware modeling for contrast enhancement and structural preservation.

To evaluate the proposed multi-objective loss, we retain $L_{{SSIM}}$  and $L_{{CON}}$and sequentially remove the other loss terms to form four comparative settings. The results in Table \ref{table6} show that the full loss achieves the best performance in SD, MS-SSIM,  $Q_{MI}$, and $Q^{ab/f}$, while also remaining competitive in SSIM and VIF. This confirms that the complete loss design is more effective in jointly preserving contrast, structure, and informative texture details.

\subsection{Computational efficiency}
To further evaluate the computational cost of the proposed method, we compare it with representative baselines under the same input resolution of 1125×938. Three indicators are reported, including inference time, FLOPs, and the number of parameters. For runtime evaluation, 10 images are randomly selected from the MSP dataset, and the average inference time is used as the reference. The quantitative results are summarized in Table \ref{table7}.

As shown in Table \ref{table7}, the proposed method contains 11.12 M parameters, requires 791.314 G FLOPs, and achieves an average inference time of 1.610 s per image. These results indicate that the proposed framework is not the most lightweight method among the compared approaches. In particular, SAGE achieves the best efficiency in both runtime and model size, while HoLoCo and MCAF also have clear advantages in parameter count. Nevertheless, compared with several strong baselines such as HSDS and SAMT-MEF, the proposed method achieves lower computational complexity while providing more competitive fusion performance. Although the introduction of the lightweight SwinBlock and luminance-guidance mechanism increases model complexity to some extent, these components also contribute to the gains in structural preservation, detail enhancement, and luminance consistency. Therefore, the proposed method should be regarded as a performance-oriented framework that achieves a favorable trade-off between fusion quality and computational cost, rather than a lightweight design.
\section{Conclusion}

In this paper, we propose a luminance-aware multi-scale polarization image fusion framework to address the limited feature representation capability of existing methods and their insufficient exploitation of linearly polarized details under complex illumination. The proposed method builds a luminance-guidance and attention-coordinated architecture for $DoLP$ images, where multi-level luminance priors are generated to adaptively enhance $DoLP$-sensitive regions, and luminance-normalized feature concatenation is introduced in the decoder for dynamic photometric modulation. By further combining local and global attention mechanisms, the proposed framework effectively improves polarization image fusion in complex scenes while maintaining a reasonable balance between performance and computational cost. To further optimize the fusion results, a multi-objective joint loss function is also designed.

In addition, we construct a new multi-scene polarization dataset, MSP, which contains 1,000 high-resolution samples collected from 17 indoor and outdoor complex lighting scenarios. Quantitative results on MSP show that the proposed method achieves leading performance in key metrics such as SSIM, MS-SSIM, $Q^{ab/f}$ and SD, while the ablation experiments further verify the effectiveness of the proposed module design, especially the contribution of the luminance-guidance mechanism and CBAM. These results demonstrate that the proposed framework provides a competitive solution for fusion  of polarization images in complex lighting environments and has potential for applications such as military reconnaissance and intelligent driving. In future work, we will further expand the dataset to include more polarization-sensitive scenarios and investigate polarization image fusion under more extreme interference conditions.

\section{Acknowledgements}
This research was supported by the Natural Science Foundation of Guangdong Province (No. 2024A1515011880), the Basic and Applied Basic Research of Guangdong Province (No. 2023A1515140077),the National NaTural Science Foundation of China (No. 52374166), 
the Research Fund of Guangdong-HongKong-Macao Joint Laboratory for Intelligent Micro-Nano Optoelectronic Technology (No. 2020B121
2030010), and the Yunnan Fundamental Research Projects (202301AV070004, 202501AS070123).

\printcredits

\bibliographystyle{cas-model2-names}

\bibliography{cas-refs}


\end{document}